# Binary Image Features Proposed to Empower Computer Vision

Soumi Ray[1], Vinod Kumar[2]

*Biomedical Engineering Laboratory, Department of Electrical Engineering, Indian Institute of Technology Roorkee, 247667, Haridwar, Uttrakhand, India*

[1]Corresponding author, Email: soumiray.eng@gmail.com, Phone number: +919758053137
[2]Vinod Kumar, Email: vinodfee@gmail.com, Phone number: +919412074172

**Abstract**

This literature has proposed three fast and easy computable image features to improve computer vision by offering more human-like vision power. These features are not based on image pixels' absolute or relative intensity; neither based on shape or colour. So, no complex pixel by pixel calculation is required. For human eyes, pixel by pixel calculation is like seeing an image with maximum zoom which is done only when a higher level of details is required. Normally, first we look at an image to get an overall idea about it to know whether it deserves further investigation or not. This capacity of getting an idea 'at a glance' is analysed and three basic features are proposed to empower computer vision. Potential of proposed features is tested and established through different medical dataset. Achieved accuracy in classification demonstrates possibilities and potential of the use of the proposed features in image processing.

1. **Introduction**

Image analysis is used in space research, real-time information tracking, security enhancement, medical imaging and in many other advance applications. The importance of image analysis lies in accurate information extraction from an image to offer more reliable details. Higher details increase the accuracy in information interpretation. Several types of features can be extracted form an image through image analysis. Those features are potential information to understand an image and its uniqueness. First order features like mean, standard deviation, variance, skewness, median, kurtosis etc. which are mainly extracted from histogram[1, 2], are often used for image classification. Haralick proposed a higher order method to get texture information based on co-relation with neighbours [3]. These higher order texture features increase accuracy of interpretation. In Haralick's method, basically pattern of image surface is extracted. To make computer vision highly accurate and reliable, researchers' primary target is extracting as many useful features as possible from an image. Till date the extracted features state almost all information about a pixel's independent and relative intensity information, along with the image's absolute and relative intensity distribution to identify its pattern [4].

To offer computer a more human like vision we need to understand human vision system. Instead of discussing in depth biological details, a rough outline of human vision system is given here. Human eyes take image information from surrounding and interpret it to make a sense about the scene. Eyes work like convex lens through which rays pass and create inverted image in focus plane [5]. Our brain nerves then analyse this image and help us to understand. So, basically human eyes read an image of a scene from a distance. Distance varies depending on the type of the scene. During reading of books, newspapers or similar thing distance will be less than a foot. For landscape it can be anywhere from few feet to infinity (for sky, stars, sun, moon etc.). In all the cases, the inverted images are created in the focus plane of eyes. When we look into any scene, we gather an overall idea about the scenario first, and then observe the details in depth depending on our requirements and interests[6, 7]. On contrary, a computer always read an image like reading a datasheet. It reads out each and every pixel sequentially like a scanning machine to gather complete details at first. Then required information is extracted from those details.

This difference in observation style sometimes put an unwanted computational load on computer during image analysis. Machine vision also lacks the feature of human vision called 'glance'. In this literature, rather than focusing on fine details of image features like colour, brightness, size, shape, boundary, texture, some new image features and their extraction methods are proposed to enhance computer vision for extraction of quick initial facts from an image with low computation load. These features will help to gather overall average information like a quick view of human eyes.

## 2. Literature review

Image is colour and intensity based two-dimensional (2D) pictorial information display of a three-dimensional (3D) scenario. These are captured and stored by common people to preserve memories and by scientists, doctors, engineers and many others as part of the technical requirements of their professions. Those requirements primarily include image analysis for further utilisation of information stored in the images. Though images can be digital or analogue, in this literature our discussion will be limited to digital images. Starting from here, the word 'image' will refer to digitally saved image only.

A digital image is nothing but a collection of pixels. For computation, it is represented by computer as 2D array, elements of which contain, mainly, intensity information of respective pixels. Number of pixels in an image varies with the size and resolution of that image. A large size, high resolution image requires higher number of pixels to store all the information. Pixels are the smallest detectable unit of a digital display.

Most prominent characteristic of an image is its colour [8, 9]. Based on the colour of an image, -it can be categorised as one of the four major types, binary image, gray image, colour image and multispectral image [10]. Binary, gray and colour images are visible in bare eyes and easy to understand. Multispectral images are capture through electromagnetic spectrums beyond our perceptual range or sonar (Sound Navigation and Ranging) system. The captured information is then mapped into the visible spectrum to enable inspection by human. Digital colour world has several colour spaces like RGB (red, green and blue), CIE (International commission on illumination), HSV (hue, saturation and brightness) [9, 11] etc. Multispectral images are mapped into any of these colour spaces. During mapping loss of information happen if the actual image contains higher number of spectral bands than the destination space.

Other than colour, texture and shape are two major properties which are useful to understand, analysis and classify an images [12-15]. Several valuable features are extracted from all three above mentioned properties of an image to enhance digital vision of a computer or robot.

All available image features can be classified as low level features and high level features based on the type of mathematical models used for feature extraction [16]. Features directly calculated from image array using array elements' value are classified as low level features. High level features are calculated from low level features or from a secondary array computed from the image array. Another popular way of feature classification is based on the target area of an image which is used in extraction process. Here, features are classified in three different categories, pixel level features, local features and global features [16]. Pixel level features are calculated directly from the independent value of image array elements. It can be intensity, colour information, location etc. Local features are extracted from region of interest i.e. keypoints of an image, whereas, for calculation of global features, entire image is taken into account [17, 18]. Some well-known local feature descriptors are Scale Invariant Feature Transform (SIFT) [19], Speeded Up Robust Features (SURF) [20], Local Binary Patterns ( LBP) [21, 22], Binary Robust Invariant Scalable Keypoints (BRISK) [23], Maximally stable extremal regions (MSER) [24], Fast Retina Keypoint (FREAK) [25], LDT [26], newly introduced LTrP [27]. Commonly used global features are image mean, image moments, image histogram, texture histogram [28] etc. These features become local when extracted only from target location of an image to survey purpose like object detection, image matching, image stitching. Features like histogram [1, 29], moments [30-32], color coherence vector (CCV) [33], color correlogram [34], Gray level cooccurance matrix (GLCM) [3], gray level run length matrix (GLRM) [35], modified color motif co-occurrence matrix (MCMCM) [36] etc have significant contribution in colour and gray image processing.

Shape features offer an idea about the image or object shape and size [37, 38]. Two different types of shape features can be calculated. One is contour based, using which boundary features can be calculated, and another one is region based. In region based technique, the pixels within that shape area are considered for extraction of different features [37].

Tactile qualities of image surface are depicted by texture features of an image [39]. Depending on extraction technique, a texture feature can be structural, statistical, model based or transform feature [40]. Though these features are equally potential for all kinds of images, a significant use of texture features are observed in gray image analysis and classification. In medical imaging field, non-contrast multispectral images are majorly mapped into gray scale images and their texture features are extracted for further analysis and disease detection [41]. Texture and shape features play an important role for disease classification too; because, in maximum cases, the target object i.e. the disease patch in the image changes its shape and size with time. Statistical texture features are further classified into three categories based on their computation processes, first order, second order and higher order [42, 43]. First order features are calculated directly from pixel values like low level features, second order and higher order are calculated from relative pixel values, that means, value of neighbour pixels are taken into account during feature calculation. Higher order texture features offer more details about an image like relative pixel intensity, patterns etc. but first order features are good for basic idea about the image. From the above discussion it is clear that there are large numbers of features available for in depth analysis of an image. Simple to complex mathematical calculations are involved in extraction of all these features. Pixel based and local features offer better information about part of an image, whereas global features offer better overall idea about the entire image under test. But till the glimpse capacity of digital vision is somehow lacking, though some significant efforts are made towards this direction by finding object boundary, coarseness and different special relationship features [6, 7, 17, 18, 44].

With advancement of AI expectations have increased. We expect computer to reflect human perception more accurately in its vision. Human brain analyses an image first by overall condition of the image[6]. The facts observed are mainly the amount of useful information in an image, image quality i.e. distorted or not and the information distribution. Colour, contrast, brightness, identification of objects etc. are being noticed subsequently. It is also observed that global features are very potential in scene perception[18]. For example, an image can be compact like image in figure 1(a) or may have information distributed in a pattern like figure 1(c) or in an unorganised way as shown in figure 1(b). If any part of noticeable area is missing in an image as shown in figure 1(d) that must be detected at first glance by human being. An idea about all these features can be assessed by calculating and combining different higher order texture features like contrast, correlation, energy, entropy etc.; but a direct measurement of distribution of information will offer a better human like 'at a glance' fast observation power to computer vision.

In the following sections of the article, we have proposed three features to provide computer a fast 'over all' idea of an image. For simplicity of understanding, colour images are not considered in the discussion of this literature. Only gray scale images having intensity ranging from 0 to 255 are studied. So, numerical value of array elements will remain bounded between 0 and 255. After discussing the basic concepts of proposed features, methods of feature calculation from an image are elaborated and followed by the application, mainly in medical image analysis. Qualitative analysis of the results is demonstrated through confusion matrix. In the discussion section, characteristics of each feature are investigated. Details definitions with formulas, range of feature values, programing algorithms for implementation of these features to develop CAD (Computer Aided Diagnostic) and enhance computer vision are given in the appendices at the end of the article.

## 3. Proposed Features

Any image, technically, has two parts- foreground and background. The foreground contains image information which is useful for further analysis. Rest of the information collectively contributes in formation of image background. Foreground can be continuous as shown in figure 1(a) and (b) or scattered over the area as shown in figure1(c). There is an unexpected loss of image information in the image shown in figure 1(d). Generally image analysis is done by extracting intensity information of each pixel. In this article, instead of each pixel, the groups of pixels are considered for feature extraction. The entire image is converted into two groups of pixels, foreground and background, using a suitable threshold value. Three major features namely information packing factor (IPF), compactness (C), porousness (P), and two minor features namely scatterness (S), total pore area (w), are proposed to get a quick idea about the image under test.

3.1 Information Packing Factor (IPF): Fraction of area of an image that is occupied by information which are useful for further analysis to extract data is proposed as Information Packing Factor [IPF] of the image.

In simple words, IPF is the measure of available foreground information in an image. In crystallography, there is a term called atomic packing factor (APF) which defines the fraction of volume of a crystal that is occupied by atoms [45]. Using the same concept, information packing factor is proposed here to describe the foreground image information amount with respect to the total image size. For a (nXm) image, I, where n is the number of rows and m is the number of columns of the image array, if total u number of pixels belong to foreground and z number of pixels belong to background, then

Information Packing Factor (*IPF*) = u / (n X m)

And $n \times m = u + z$

So, $IPF = u/u+z = 1/(1 + z/u)$     (1)

Now the value of IPF varies depending on the values of u and z as follows

$$IPF \begin{cases} = 1 & \text{If } z=0 \text{ and } u>0; \text{ no background pixel in the image} \\ < 1 & \text{If } z>0 \text{ and } u>0 \\ = 0.5 & \text{If } z=u \\ = 0 & \text{If } u=0 \end{cases}$$

Hence, theoretically the range of possible values of IPF can be written as 0≤*IPF*≤1. But, IPF becomes 0 when *u*=0. And when *u*=0 then the image contains no information to be processed. So, in practice, *u*=0 is not possible. When *z*=0, *IPF*=1 i.e. there is no background in the image. When *z*=*u*, *IPF*=0.5 as per equation 1. Henceforth, practical the range can be written as 0<*IPF*≤1. It is also observed that the value of IPF is inversely proporsonal to the value of *z*. This first order observation will offer an overall idea about the amount of available foreground information in an image.

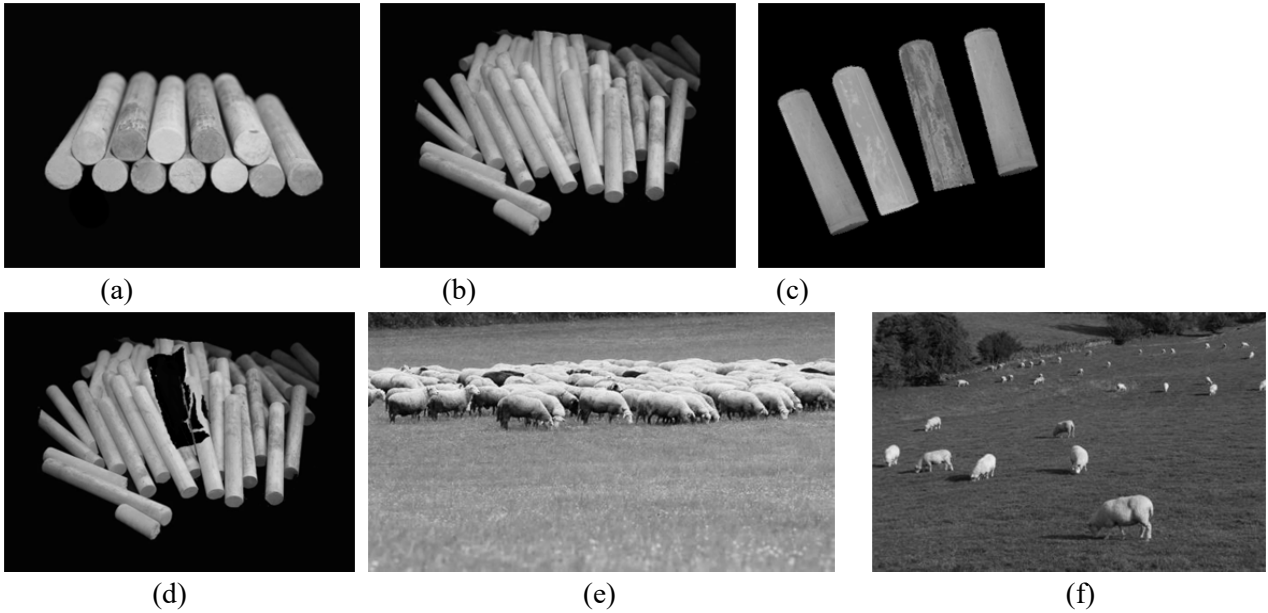

*Figure 1: (a) & (b) compact information, (c) scattered information, (d) unexpected gaps in information [46], (e) cattle are in a group [47], (f) cattle are scattered over the field [48]*

3.2   Information Compactness (C): Fraction of information spread area of an image that is actually occupied by foreground pixels is described as compactness [C] of the image.

Compactness offers the idea that how closely the usable information is packed in the image. It can also be described as tightness of foreground information. The measurement is local as only the spread area is counted not the area of whole image.

It is another quick observation of human eyes is the spread of recognisable things within the total available area of image. Images without and with gaps in information distribution are shown in figure 1. Figure 1(a) has highly compact information. No significant gap is there in the image information distribution. Such arrangements offer higher compactness value. Figure 1(b)-(d) have one or multiple area(s) where there is (are) gap(s) between information. With increase in gaps, compactness value decreases. We can see, cattle are grazing in the field in figure 1(e) and 1(f). Our eyes can easily recognize the difference in information pattern. In figure 1(e) cattle are forming a group whereas in figure 1(f) they are scattered over the field. Here we can consider the field as image background and cattle as foreground information. If all foreground information is placed side by side without any gap, it can be considered as completely compact. The highest compactness value should be 1. If u number of foreground pixels are there in (n X m) image, and y number of background pixels are placed within the foreground information distribution, then compactness is the ratio of image information to total area covered by foreground information and in-between background pixels.

$Compactness\ (C) = u/(u+y) = 1/(1+y/u)$

Now the value of C varies with the value of u and y.

$$C \begin{cases} = 1 & \text{If } y=0 \text{ and } u>0; \text{ no space between foreground pixels} \\ < 1 & \text{If } y>0 \text{ and } u>0 \\ = 0 & \text{If } u=0 \end{cases}$$

So, theoretically the range of the value of *C* can be written as $0 \leq C \leq 1$. But in practice, *u*=0 is not possible. So, *C* can't be 0. With no gap between information i.e. *y*=0, compactness turns to 1. So, practically the range can be written as $0<C\leq1$. As y increases with constant u, the numerical value of compactness decreases. The maximum value of y can be *z*. so the range of y can be written as $0\leq y\leq z$.

Another characteristic which is just opposite of compactness can be measured by deducting compactness value form unity i.e. the maximum possible compactness. It can also be measured by directly taking ratio of in-between background pixels to the total area covered by the same and foreground information pixels. This characteristic is addressed as scatterness in this literature.

$Scatterness\ (S) = \bar{C} = 1 - C$

$$= 1 - u/(u+y) = y/(u+y) = 1/(1+u/y)$$

Now the value of *S* varies depending on the values of *u* and *y*.

$$S \begin{cases} = 1 & \text{If } y>0 \text{ and } u=0 \\ < 1 & \text{If } z>0 \text{ and } u>0 \\ = 0 & \text{If } y=0;\ no\ space\ between\ foreground\ pixels. \end{cases}$$

So, the theoretical range of possible values of S can be written as $0\leq S\leq1$. In practice, $u=0$ is not possible, so, practically the range can be written as $0\leq S<1$. The maximum value of y can be the value of *z*. So the range of y can be defined as $0\leq y\leq z$.

Scatterness (S) can be defined as fraction of information spread area of an image that is occupied by non-usable or background pixels. Scatterness offers the idea that how loosely the usable information is placed in the image.

3.3 Image Porousness (P): Fraction of information spread area of an image that is occupied by non-usable information which has no link to the background pixels as neighbor is described as porousness [P] of the image.

Missing information confined within foreground information and completely isolated from background contributes to porousness of an image. Each isolated area creates a 'pore' in the image. Porousness can be created due to several reasons. It can be due to the nature of the image. It can be because of unfortunate deletion of information of an already saved image. Generally such pores (the background parts trapped within foreground) are filled by background intensity value which is most commonly 0 or 255 in case of gray images.

All pores are made of scatter pixels but each scatter pixel may not be part of a pore. It is sometimes tough to differentiate porousness from scatterness in bare eyes. In both the cases, there are background pixels in between foreground pixels. In case of scatterness the gaps are created due to unattached spread of foreground information. Porousness can be identified as the non-peripheral part of scatterness. Pores should have no link with background of the image. It needs be completely confined within foreground. To explain, an example of an image of islands in a sea shown in figure 2(a) is taken. In this image, lands are not connected to each other but they are complete on their own. If sea is considered as background then the gaps between islands, due to sea, lower the compactness of the image. Now, say, in one of these islands there is a large lake which has no link to the actual sea. Then this lake will contribute to porousness, because a land itself is expected to be continuous. As shown in figure 2(b), unexpected low intensity volcanic crater in a landscape has introduced porousness in that image.

Porousness is measured as ratio of background pixels belongs to pores to the sum of foreground information pixels and total pore pixels. When total area occupied by pores is w, then

Porousness $(P) = w/(u+w)$,    $w\leq y\leq z$.

$$= 1/(1+u/w)$$

Now the value of P varies with the values of *u* and *w*.

$$P \begin{cases} = 1 & \text{If } u = 0 \\ < 1 & \text{If } w > 0 \text{ and } u > 0 \\ = 0 & \text{If } w = 0; \quad \text{no pore in the image.} \end{cases}$$

So, theoretically the range of possible values of P can be written as $0 \le P \le 1$. In practice, as *u*=0 is not possible the range can be written as $0 \le P < 1$. The maximum value of w can be y and maximum value of y can be z. So the range for w can be written as $0 \le w \le y \le z$.

### 3.3.1 Advanced analysis of porousness

As discussed above, pores remain confined within the image foreground area without having any link to the image background. The count of such pores offers the number of porous parts. When *w*>1, the total porous area can be divided into one or several pores. Each pore contributes to the porous area of an image. Using any established connected area labelling method [49, 50], pore count and volume measurement of each pore can be done.

For an example, consider there are *q* pores in an image which has u numbers of foreground pixel and pixel count of the pores are $w_1, w_2, \ldots w_q$ respectively, then

$$\text{Total pore area } w = \sum_{i=1}^{q} w_i$$

$$\text{Pore count } (n_p) = q$$

And, porousness $P_i$ of $i^{th}$ pore is
$P_i = w_i/(u + \sum w_j)$ where j = 1 to q
$= w_i/(u + w)$

So, $P = \sum P_i$ and the range of $P_i$ is same as P.

Sometimes, very fine gaps in compact images also offer porousness. Volume of such pores normally remains very low. This fact has been discussed later in this article.

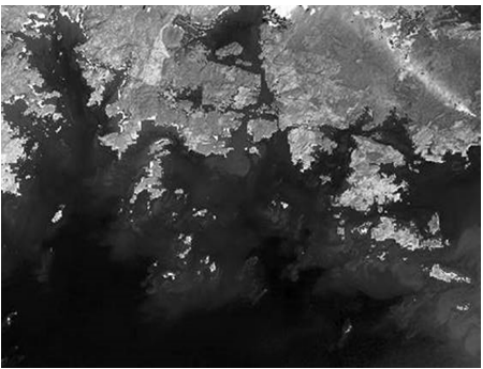 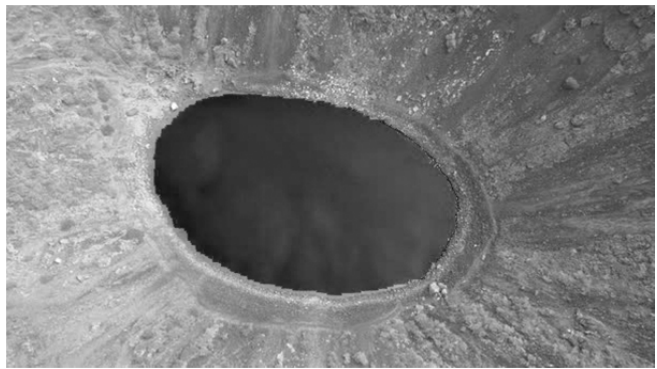

(a)            (b)

*Figure 2: (a) land and islands[51] (b) land with volcanic crater[52]*

## 4. Feature Extraction

### 4.1 Pre-processing of images

In all above mentioned features, images are initially considered as combination of two levels – foreground and background. This assumption leads to binarization of an image. One level contains background information and second level contains foreground information. For a gray scale image, in most of the cases, background colour remains black or white. Otherwise, it can be user defined. And any other intensity level beyond background intensity range is considered as foreground information.

This binarization can be done by user defined threshold value. Depending on user's understanding, a particular intensity value or a range can be selected to divide the image into two pixel groups for further analysis. To do thresholding automatically, any established binary thresholding method [53, 54] can also be used. In this article, Otsu bi-level thresholding is used to convert the gray images to binary images. The threshold calculated by software is mentioned for each case.

### 4.2 Information extraction

For computation, a binary image is mathematically represented by an array of two elements. To keep it simple we have considered 0 and 1 as array elements. 0 presents the background and 1 represents the foreground elements. An image array (I) is considered to describe the feature calculation process from an image. The pictorial demonstration is shown in figure 3(a). In this image array, all 0s are representing background information (BI) location and 1s are presenting foreground information (FI) location. Total length of foreground information spread (L) for each row is calculated from the image array and presented in 'foreground information length' array. Similarly, count of available foreground information per row and background pixels within that are calculated and presented in 'foreground information count' and 'scatterness count' array. From scatterness count array, pore pixels are extracted. The scattered pixels having no link to background pixel as neighbour are considered as pore pixels. For neighbour search 8-connected neighbourhood are taken into account. The details of extraction are highlighted for row3 (R3) of image array. The scattered pixels of R5 and R6 are not contributing in pores. 2nd scattered pixel of R6 has a background neighbour. This particular scattered pixel is neighbour of another scattered pixel of R6 and the scattered pixel of R5. So, none of these three has any contribution in porousness.

A tabulation chart can be formed from image array for easy calculation of the features. The formation of tabulation chart is presented in figure 3(b). This chart provides all information required to calculate values of IPF, C, S and P.

Images, given in figure 1 (a)-(d) and 2(b), are converted into binary images as shown in figure 4 (a)-(e) respectively before further analysis. Display images have black (intensity 0) background and foreground is presented in white (intensity 255). Foreground information intensity value 255 is selected purposefully, for display only, as 0 and 1 offer no visible significant difference in digital gray images. Binarization is done by Otsu thresholding. Threshold values selected by Otsu method are 86, 80, 72, 78, 148 for figure 4(a) to 4(e) respectively.

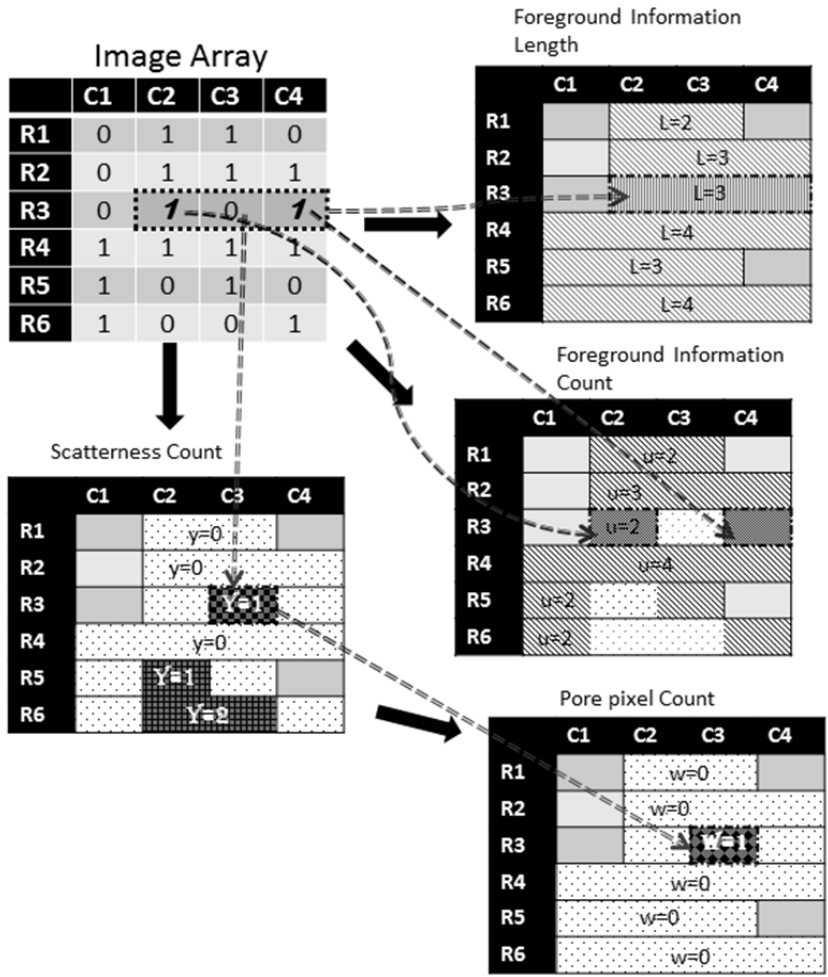

(a)

(b)

*Figure 3: (a) Information extraction process, (b) feature tabulation chart*

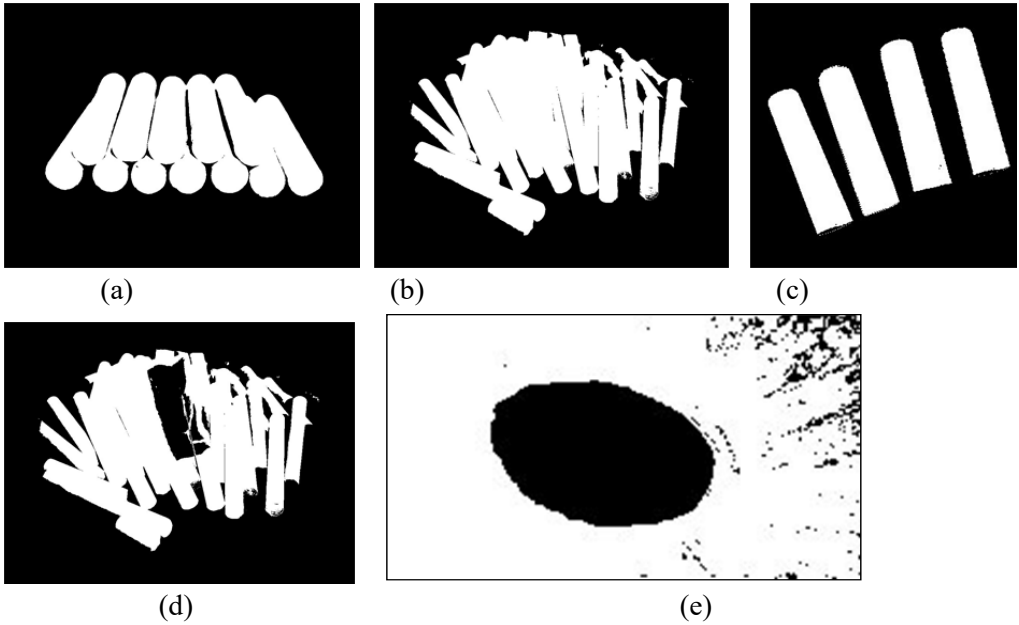

*Figure 4: (a)-(b) compact information, (c) scattered information (d)-(e) porous information*

Table 1 shows the IPF, C and P calculation of figure 4(a)-(e). Human visual perception at a glance is reflected in this calculation. As per human vision figure 1(b) has highest density of information among first four images. This fact is satisfied by calculated values of IPF. IPF value is maximum for 4(b). Human perception for figure1(a) as most compact and 1(c) as most scattered is also reflecting in calculation of proposed feature as shown in table1. Porousness is highest in figure 1(d) among first four images and it is maximum for figure 1(e) among all. Figure 1(c) has almost no porousness. Thus all the calculated features are completely satisfying human visual perception.

Area and porousness of each pore can be calculated for every single image. In table 2, complete results for figure 4(a) and 4(c) and partial results (11 for each) of figure 4(b), 4(d), 4(e) are presented in descending order. Rest parts are avoided due to the presence of huge number of very small pores.

The result shows that fine gaps in foreground in a compact image or small gaps due to scatterness sometimes unexpectedly contribute in porousness measurement. But those volumes are too low to be considered as shown in table 2. Here only figure 4(d) and 4(c) have significant pores, one for each, which are emphasized by bold font in table. The threshold for considerable pore volume must be application specific and can be determined by end users.

| Image | Image size nXm | Info. Size p | Scatter size q | Total pore area w | Pore count $n_p$ | IPF | C | P |
|---|---|---|---|---|---|---|---|---|
| Fig.4(a) | 786432 | 197719 | 27484 | 3038 | 29 | 0.2514 | 0.8780 | 0.0151 |
| Fig.4(b) | 786432 | 274513 | 46127 | 5871 | 77 | 0.3491 | 0.8561 | 0.0209 |
| Fig.4(c) | 122500 | 36697 | 17364 | 8 | 4 | 0.2996 | 0.6788 | 0.0002 |
| Fig.4(d) | 786432 | 248985 | 72636 | 32616 | 62 | 0.3166 | 0.7742 | 0.1158 |
| Fig.4(e) | 26015 | 20041 | 5895 | 5463 | 114 | 0.7704 | 0.7727 | 0.2142 |

Table 1: IPF, compactness and porousness measure

| Image | Pore count $n_p$ | Pore area per pore (pixel count) | % Porousness per pore | Image | Pore count $n_p$ | Pore area per pore (pixel count) | % Porousness per pore |
|---|---|---|---|---|---|---|---|
| Fig.4(a) | 29 | 2011 | 1.0020 | Fig.4(b) | 77 | 1473 | 0.5254 |
| | | 385 | 0.1918 | | | 1137 | 0.4055 |
| | | 317 | 0.1579 | | | 827 | 0.295 |
| | | 141 | 0.0702 | | | 527 | 0.188 |
| | | 36 | 0.0179 | | | 500 | 0.1783 |
| | | 32 | 0.0159 | | | 320 | 0.1141 |
| | | 21 | 0.0105 | | | 304 | 0.1084 |
| | | 19 | 0.0095 | | | 180 | 0.0642 |
| | | 15 | 0.0075 | | | 159 | 0.05671 |
| | | 14 | 0.0070 | | | 95 | 0.03388 |
| | | 5 | 0.0025 | | | 83 | 0.0296 |
| | | 5 | 0.0025 | Fig.4(d) | 62 | **25539** | **9.069** |
| | | 5 | 0.0025 | | | 1469 | 0.5217 |
| | | 5 | 0.0025 | | | 1144 | 0.4062 |
| | | 4 | 0.0020 | | | 1084 | 0.3849 |
| | | 3 | 0.0015 | | | 954 | 0.3388 |
| | | 3 | 0.0015 | | | 773 | 0.2745 |
| | | 3 | 0.0015 | | | 499 | 0.1772 |
| | | 2 | 0.0010 | | | 479 | 0.1701 |
| | | 2 | 0.0010 | | | 311 | 0.1104 |
| | | 2 | 0.0010 | | | 151 | 0.05362 |
| | | 1 | 0.0005 | | | 46 | 0.01634 |
| | | 1 | 0.0005 | Fig.4(e) | 114 | **5190** | **20.35** |
| | | 1 | 0.0005 | | | 27 | 0.1059 |
| | | 1 | 0.0005 | | | 20 | 0.07842 |
| | | 1 | 0.0005 | | | 15 | 0.05881 |
| | | 1 | 0.0005 | | | 7 | 0.02745 |
| | | 1 | 0.0005 | | | 7 | 0.02745 |
| | | 1 | 0.0005 | | | 6 | 0.02353 |
| Fig.4(c) | 4 | 3 | 0.0082 | | | 6 | 0.02353 |
| | | 2 | 0.0054 | | | 6 | 0.02353 |
| | | 2 | 0.0054 | | | 6 | 0.02353 |
| | | 1 | 0.0027 | | | 6 | 0.02353 |

Table 2: In-depth calculation of porousness for each pore

### 5. Application of Features in Image Analysis

Using these features, primary classification of a set of images can be done to reduce complexity and computational load on advanced steps of analysis. IPF presents volume of available information in an image. This is very basic but required feature. Classification based on information size can be powered by IPF. Compactness (also scatterness) and porousness illustrate distribution of foreground information within image. Porousness pixels are also counted in scatter counting whereas all scattered pixels may not be considered during porousness calculation. All these features are extracted based on the level of individual pixel intensity with respect to threshold, keeping the computation fast and light weight.

In several applications like medical image processing, satellite image analysis, remote sensing, robotic vision, document processing, automatic inspection etc. feature extraction is required for image analysis, processing and decision making. Knowledge about information density and distribution will help in almost all the cases to understand images in a better way.

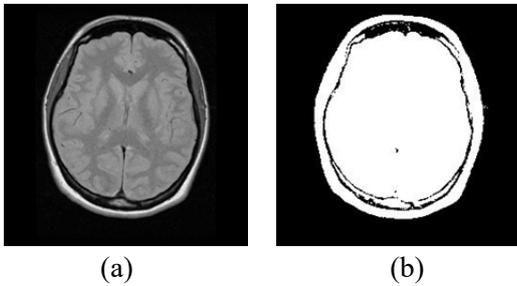

(a)                 (b)

*Figure 5: (a) Brain MR scan image[56] and (b) its binary equivalent image*

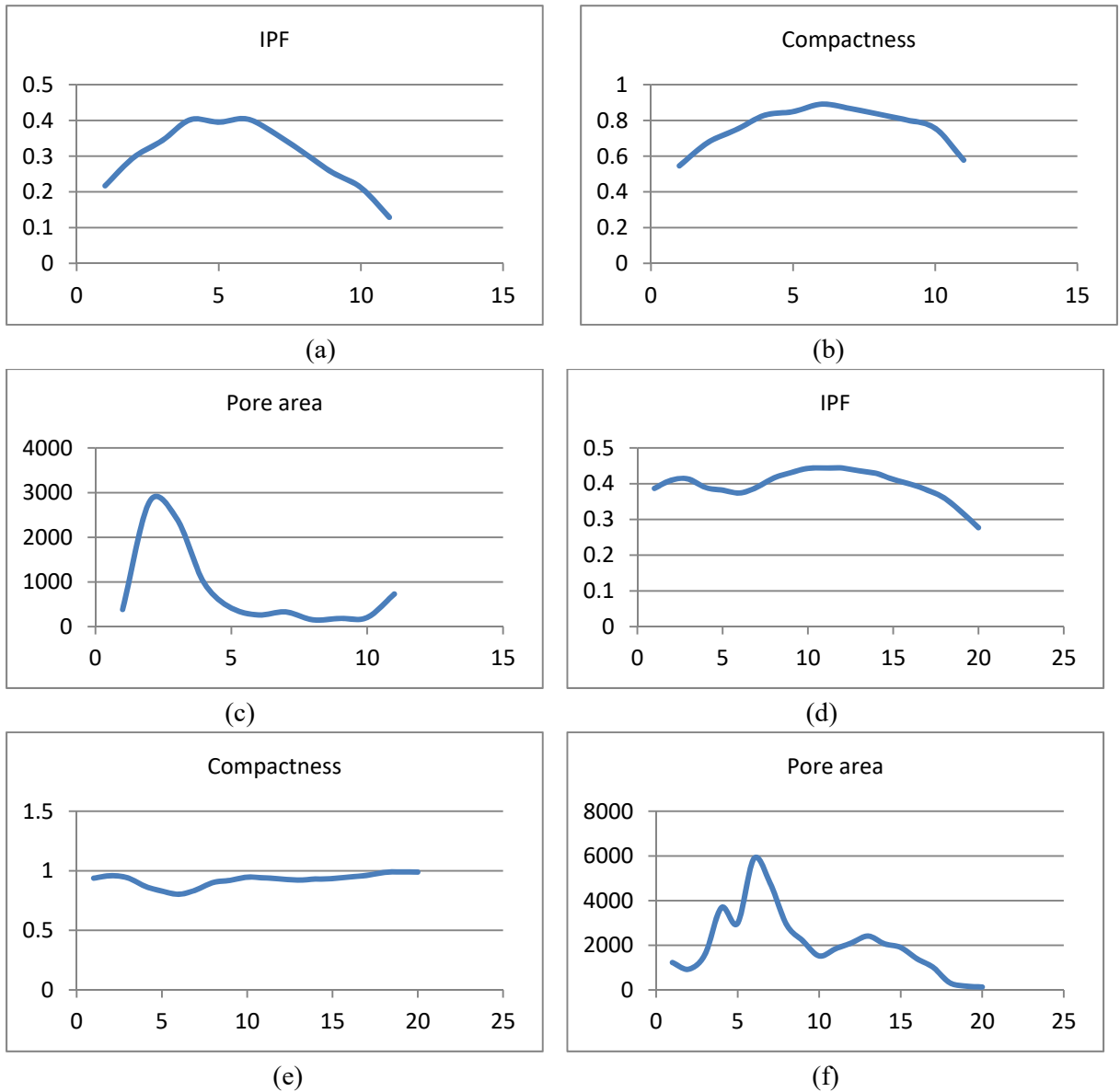

*Figure 6: (a)-(c) features plot for MR dataset 1 and (d)-(f) features plot for MR dataset 2*

In brain diseases like dementia, Alzheimer brain size starts shrinking over time. A longitudinal brain scan also offers change in brain size from slice to slice. Using impact of this fact, scanned slices can be classified primarily by IPF. In a transverse multi slice brain scan, brain area in different slice is different. At the base of skull, eyes are get scanned and the total brain area remains low. Scanning starts from this level and gradually moves towards the top of the head and vice versa. Approximately at the middle of the scanning range, lateral ventricle level of brain is scanned. Scanned slices are automatically saved in the scanning order by computer. Close observation shows that this order of scans offers a particular pattern to IPF, compactness and porousness volumes. Two healthy brain transverse MR scan dataset are collected from open-source internet database [55], [56] to perform a test. Background threshold values are selected by Otsu method. One scan image and its binary equivalent are shown in figure 5(a) and 5(b) respectively. The variation of IPF, C and w of a MR dataset are shown in figure 6(a)-(c) and the same of another in figure 6(d)-(f).

Another dataset of 32 slice brain CT, collected from internet archive site[57], is used for further investigation. The collected dataset is pre-processed to remove the skull. The image of one skull removed CT scan and its binary equivalent is shown in figure 7(a) and (b). Skull removed scans are taken to extract IPF, compactness and porousness. Extracted values are plotted and shown in figure 8(a)-(c). The main difference in MR and skull removed CT images is that the intensity value for lateral ventricle is high for MR and fall in the range of background for CT.

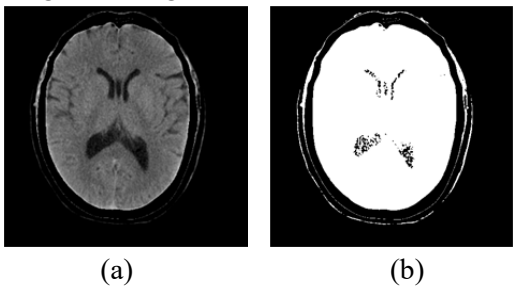

(a)             (b)

*Figure 7: (a) Skull removed brain CT scan[57] and (b) its equivalent image*

IPF and compactness remain always higher for lateral ventricle level brain slices where largest area of brain is captured during scanning. Porousness is high for slices at the base of the skull where eyes are included in scan and lower for the rest. In this CT dataset the last slice, displayed in figure 9(a), is showing very high porousness. The binary equivalent of skull removed image of figure 9(a) has no brain but background trapped within a circular brain like patch as shown in figure 9 (b). Any such very high value thus will present error in data of that particular CT slice. For further analysis of porousness, average pore area of CT dataset is calculated by using the formula stated in equation 2. The plot of the same is shown in figure 8(d). Error is amplified in this plot and very easily recognizable.

Average pore area = $w_{avg} = w/n_p$                                                                                                            (2)

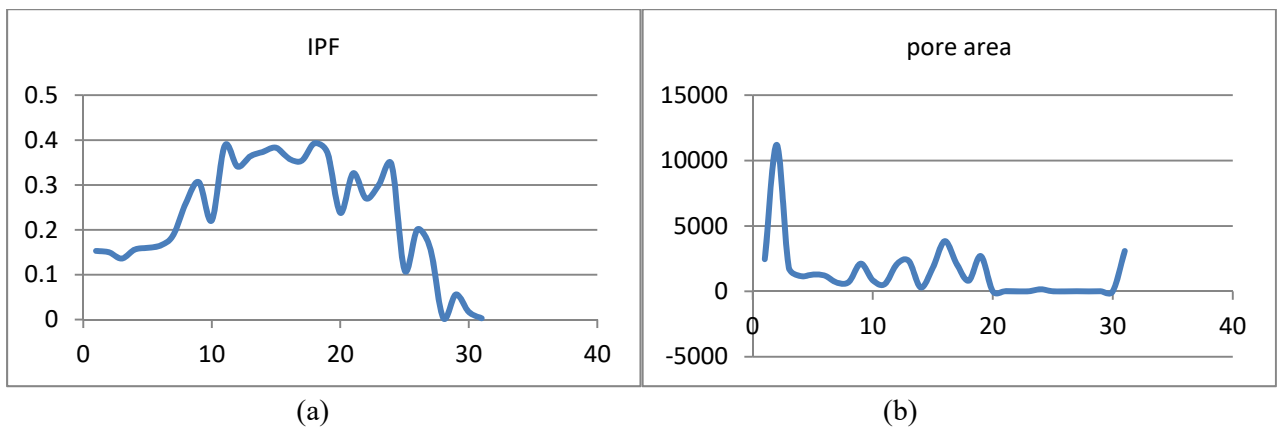

(a)                                                (b)

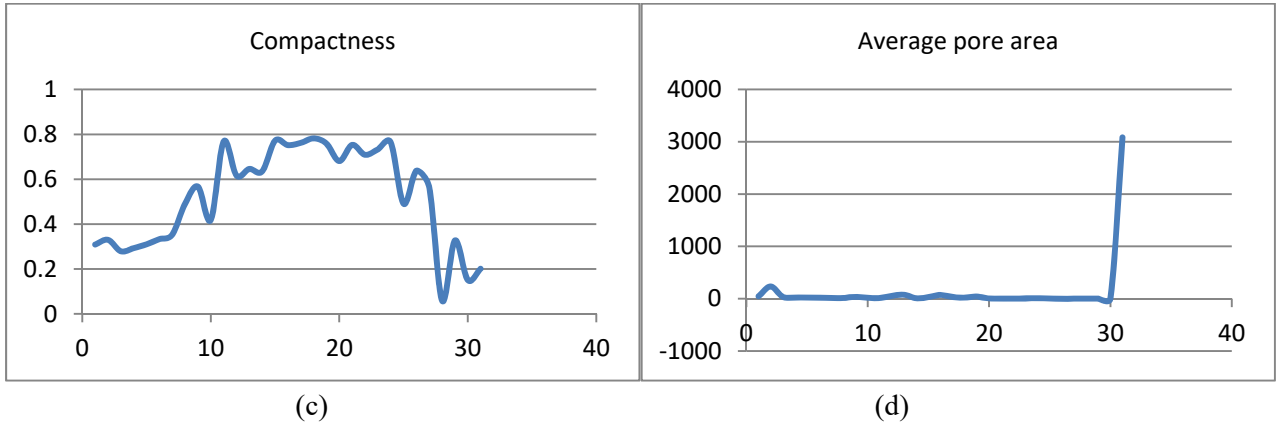

Figure 8: features plot for CT dataset

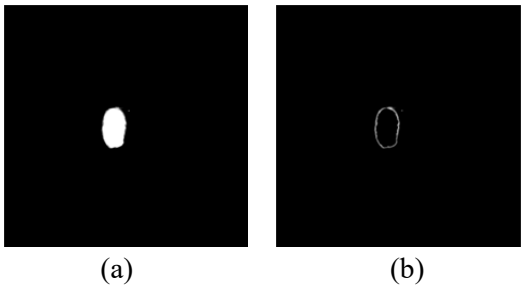

(a)          (b)

Figure 9: (a) faulty brain scan[57] (b) skull removed image

To explore the potential of the proposed features- IPF, compactness (C) and porous area (w), classification of brain CT scan dataset is performed. Total 142 CT scan slices in which 46 slices have brain as well as eyes in scan, 64 slices have brain but no eye area and 32 slices have no brain head scan as shown in figure 9(a) are used. The dataset is collected from free online data source[57]. Complete set is initially divided into two categories – slices having eyes in scan and slices without eyes. Matlab Neural Network Pattern Recognition (nprtool) toolbox is used to do classification. 55% of entire dataset is utilized to train the network of 10 hidden neuron layers. The classification accuracy of all confusion matrixes, shown in figure 10, is 97.9% whereas the test data is classified with an accuracy of 96%.

Now the dataset of second class, without eyes, is further divided into two classes – slices with brain, slices without brain. Total three classes are then used to train and test the neural network of 10 hidden neuron layers for classification. Obtained confusion matrices presented in figure 11 describes overall accuracy as 95.8%.

To classify this database in the above mentioned three classes, more features are included in different combination to observe the change in classification accuracy. Previously IPF, C and w are used. Now pore count $n_p$ and porousness P are also included in input. The different input feature combinations are as follows:-

Combination 1 [C1]: IPF, C, w

Combination 2 [C2]: IPF, C, w, $n_p$

Combination 3 [C3]: IPF, C, w, P

Combination 4 [C4]: IPF, C, w, $n_p$, P

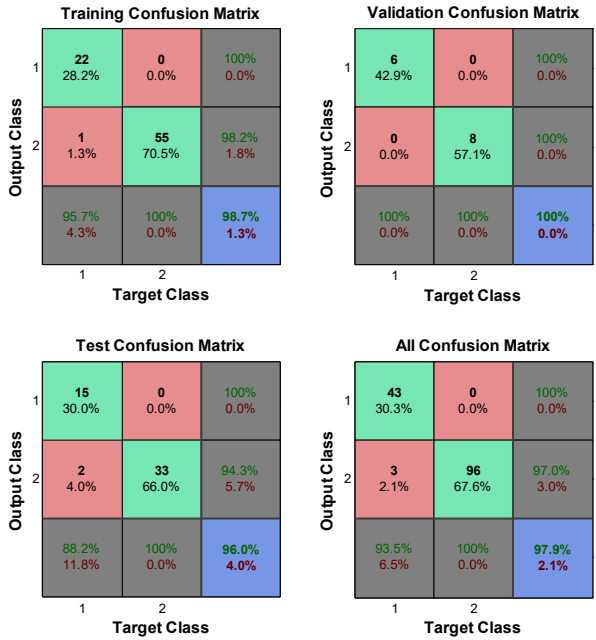

*Figure 10: confusion matrix of 2 class classification of CT scan*

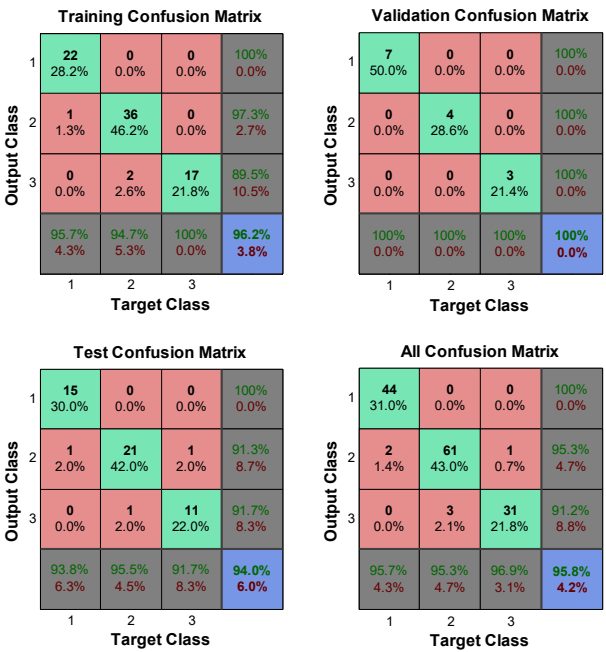

*Figure 11: confusion matrix of 3 class classification of CT scan*

Every time, randomly 55% of data is used for training and 35% of data is left for testing of the neural network of 10 hidden layers. For each combination, 10 test results are collated in table 3. The average accuracy offers an approximate overview of classification potential of each combination. The observations are as follows:

(1) Average accuracy of test data classification has been gradually increased.
(2) Performances of C3 and C4 are very similar with respect to the average accuracies.
(3) Standard deviation in test data classification accuracy is minimum for C2. C4 also has offered very close value.

(4) Maximum accuracy obtained in training set classification is 100% by C4.
(5) Maximum accuracy obtained in test set classification is 98% by C3.
(6) C3 and C4 have shown better potential in overall classification accuracy.

Total such 31 input combinations are possible from these 5 features. End users can determine the best suitable combination for their specific applications. The best performing network of that combination can be saved for further use.

|   | C1 | | | C2 | | | C3 | | | C4 | | |
|---|---|---|---|---|---|---|---|---|---|---|---|---|
|   | Training | Test | All | Training | Test | All | Training | Test | All | Training | Test | All |
| 1 | 94.9 | 90 | 93 | 87.2 | 92 | 87.3 | 97.4 | 98 | 97.9 | 97.4 | 96 | 97.2 |
| 2 | 87.2 | 68 | 80.3 | 93.6 | 88 | 91.5 | 69.2 | 70 | 68.3 | 100 | 94 | 96.5 |
| 3 | 89.7 | 78 | 85.9 | 92.3 | 82 | 89.4 | 93.6 | 96 | 95.1 | 94.9 | 86 | 92.3 |
| 4 | 85.9 | 80 | 81 | 74.4 | 76 | 74.6 | 94.9 | 86 | 91.5 | 89.7 | 84 | 88.7 |
| 5 | 66.7 | 60 | 63.4 | 73.1 | 68 | 72.5 | 98.7 | 92 | 96.5 | 69.2 | 70 | 69 |
| 6 | 96.2 | 96 | 95.8 | 74.4 | 66 | 72.5 | 70.5 | 60 | 66.9 | 96.2 | 92 | 95.1 |
| 7 | 78.2 | 70 | 75.4 | 92.3 | 88 | 90.8 | 94.9 | 96 | 95.8 | 89.7 | 82 | 87.3 |
| 8 | 84.6 | 82 | 83.8 | 97.4 | 86 | 93 | 92.3 | 90 | 91.5 | 62.8 | 72 | 67.6 |
| 9 | 94.9 | 96 | 91.5 | 89.7 | 90 | 90.1 | 91 | 76 | 83.8 | 96.2 | 92 | 93.7 |
| 10 | 91 | 72 | 84.5 | 97.4 | 80 | 91.5 | 94.9 | 96 | 95.8 | 97.4 | 94 | 95.8 |
| Average | 86.93 | 79.2 | 83.46 | 87.18 | 81.6 | 85.32 | 89.74 | 86 | 88.31 | 89.35 | 86.2 | 88.32 |
| std. dev. | 9.00 | 12.12 | 9.42 | 9.63 | 9.08 | 8.51 | 10.72 | 13.03 | 11.63 | 12.82 | 9.26 | 11.04 |

*Table 3: % accuracy of different confusion matrices of different feature combinations*

## 6. Discussion

To calculate the features, entire image is segmented into two parts based on the value of absolute intensity of each pixel in the image. After segmentation, not pixel intensity values but the labels of these two segments are used for further calculation to keep the system simple and fast. Already established image features are categorized as first order, second order or higher order depending on their extraction methods. First order features are calculated from absolute pixel intensity values. Second and higher order features are extracted from relative intensity values of pixels. Proposed methods are calculated from binary image where not even absolute intensity values are directly considered for calculation. Segmented binary levels are used for feature extraction. So, these features can be categorized as 'binary features' or following the already used pattern can be addressed as 'zero order features'.

Logical relation between proposed image features i.e. foreground part (F), scattered gaps (G), porousness parts (D) and background (B) be can demonstrated by Venn diagram as shown in figure 12. Mathematically it can be described as $B \cap F = \emptyset, B \cap G = G, G \cap D = D, D \subset G$ and $G \supset D, G \subset B$ and $B \supset G, D \subset B$ and $B \supset D$.

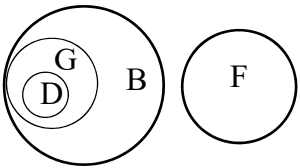

*Figure 12: conceptual presentation of image pixel intensity distribution*

Let us observe the effects of rotation of an image on the proposed features. IPF and porousness features are image orientation independent. Rotation of image has no effect in these feature values. Let us rotate the image array shown in figure 3(a) by 90 degree. The new image array I' is shown in figure 13. L, u, z, y and w from I' are extracted and compared with the same of I. The results presented in figure 13 are demonstrating no change in u, z and w. But a significant change in L and y are observed.

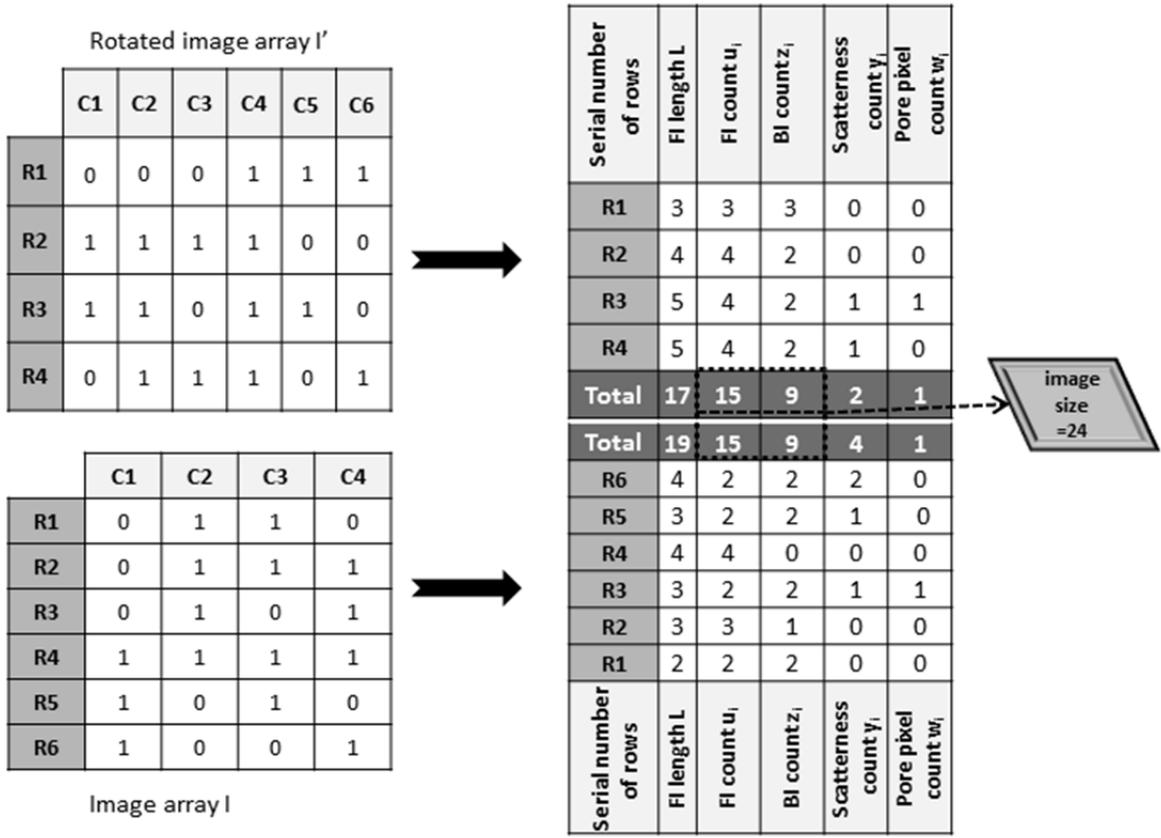

*Figure 13: Comparison of features after 90° rotation of image*

By proposed definition, IPF = u/(u+z)  where u+z = image size,

Image size retains same value after rotation. So, we can say, IPF ∞ u.

As u is not affected by rotation, IPF remains unchanged after rotation.

Porousness P is a function of u and w and both of these values remain unaffected by rotation. Hence, P also remains unaltered after rotation.

Compactness depends on u and y where y changes after rotation of image. So, C as well as S is image orientation dependent features. But if the image is just flipped i.e. the total rotation is 180 degree, no features proposed in this article will be affected.

So, the key facts of proposed features can be listed as

(1) Images under process must have some usable information.
(2) Proposed features can be addressed as 0th order binary features.
(3) Rotation of image will not affect IPF and porousness values.
(4) Compactness and scatterness are image orientation dependent features until and unless the rotation angle is 180 degree.
(5) Features, except IPF, are calculated with respect to the target area, majorly covered by foreground information. Entire image i.e. image including background area is not taken into account. So, it can be concluded that IPF is a global feature and rest are local features.

## 7. Conclusion

Few quickly computable image features are proposed in this work. These features seem to have impressive applicability in different types of image data. Proposed features are extracted from both CT and MR image of brain for result analysis. Initial outcomes are promising. Degree of accuracy can be increased using these features in combination with higher order texture features at a cost of computational load. Depending on requirement, that can be decided by user. It has been observed that without doing detail pixel by pixel intensity evaluation, fast sorting of images depending on its foreground volume and information spreading is possible. Unusual information gap in an image can be identified quickly by porousness feature. All these results can provide guidance for further in depth analysis of an image or image dataset. These features are promising for initial selection of images to reduce load on advance analysis processes. The enhancement of computer vision lies in this quick initial selection method which is intuitive in human beings. It will grab the overview first to make a basic understanding of an image.

All, except IPF, among the proposed features, are local features. For those features, during calculation of the value, entire background information is not taken into account. The proposed features are extracted from gray scale images. To extract these features from colour images additional investigations are necessary. As proposed application is developed for binary analysis, mathematical models need to be modified to handle higher order image analysis.

**Acknowledgement**

Authors would like to thank the host institute for the laboratory, internet and other essential facilities to conduct the research. This research did not receive any specific grant from funding agencies in the public, commercial, or not-for-profit sectors.

**Disclosure statement**

The authors declare no conflicts of interest.

**Funding**

This research did not receive any specific grant from funding agencies in the public, commercial, or not-for-profit sectors.